\documentclass[runningheads]{llncs}
\usepackage{graphicx}
\usepackage{float}
\usepackage{subcaption}
\usepackage[ruled, lined, longend, linesnumbered]{algorithm2e}
\usepackage{tikz,xcolor,hyperref}

\begin{document}
\title{Evolving Character-Level DenseNet Architectures using Genetic Programming}
\titlerunning{Evolving char-DenseNets using Genetic Programming}

%
%
\author{Trevor Londt\orcidID{0000-0002-9428-8089} \and
Xiaoying Gao\orcidID{0000-0002-6326-7947} \and
Peter Andreae\orcidID{0000-0002-2789-680X}}
\authorrunning{T. Londt et al.}
%
\institute{School of Engineering and Computer Science,\\
Victoria University of Wellington, Wellington, New Zealand \\
\email{\{trevor.londt, xgao, peter.andreae\}@ecs.vuw.ac.nz}\\
}
\maketitle              
\begin{abstract}
DenseNet architectures have demonstrated impressive performance in image
classification tasks, but limited research has been conducted on using
character-level DenseNet (char-DenseNet) architectures for text
classification tasks. It is not clear what DenseNet architectures are
optimal for text classification tasks. The iterative task of designing, training and testing of char-DenseNets is an NP-Hard problem that requires expert domain knowledge. Evolutionary deep learning (EDL) has been used to automatically design CNN architectures for the image classification domain, thereby mitigating the need for expert domain knowledge. This study demonstrates the first work on using EDL to evolve char-DenseNet architectures for text classification tasks. A novel genetic programming-based algorithm (GP-Dense) coupled with an indirect-encoding scheme, facilitates the evolution of performant char-DenseNet architectures. The algorithm is evaluated on two popular text datasets, and the best-evolved models are benchmarked against four current state-of-the-art character-level CNN and DenseNet models. Results indicate that the algorithm evolves performant models for both datasets that outperform two of the state-of-the-art models in terms of model accuracy and three of the state-of-the-art models in terms of parameter size.
\end{abstract}
%
%
\section{Introduction}
Natural language processing (NLP) has benefited significantly from the application of deep learning \cite{deeplearning} techniques in recent years \cite{Zhang2015}\cite{Conneau2017}\cite{Collobert2011}\cite{Yang2019}. In particular, the task of text classification has gained impressive performance increases. Long short-term memory (LSTM) \cite{LSTM} models have traditionally been the architecture of choice for text classification tasks. However, transformer \cite{devlin2018bert} models and convolutional neural networks (CNN) \cite{Zhang2015}\cite{Conneau2017}\cite{Le} outperform LSTM's for text classification tasks. While transformer models dominate the leaderboards, they have the disadvantage of containing massive numbers of parameters and the need for pre-trained language models. Only a limited number of languages have pre-trained language models available. CNN's are generally not encumbered by these limitations and have demonstrated excellent success for text classification tasks.
\par
\vspace{-0.5mm}
There are two approaches when using CNNs for text classification: word-level (word-CNN) and character-level CNNs (char-CNN). Word-level approaches generally outperform character-level approaches \cite{Zhang2015}. However, word-level approaches require a pre-trained word-model such as word2vec \cite{church_2017} or GloVe \cite{Pennington}. This requirement is a similar limitation as seen with transformer models. Further, an additional disadvantage of  word-level CNNs is that they require the input text to be pre-processed by removing stop words, stemming words, removing punctuation and dealing with out of vocabulary words. Each of these operations increases the likelihood of the input text being corrupted or important discriminating information being inadvertently removed. There is also the potential problem of not having a pre-trained word-model available for a particular language. Char-CNNs require no pre-trained word or language models. Also, char-CNNs require no pre-processing of the input text data, mitigating any potential errors introduced through incorrect pre-processing. The disadvantage of char-CNNs is that they are, in general, less accurate than word-level CNNs. Research has shown that adding depth to a char-CNN does not result in breakthrough performance improvements, as shown in the image classification domain. DenseNet \cite{Huang2016} is a network architecture that has shown good performance in the image domain and has recently shown promising results in the text classification domain \cite{Le}. However, the results have still not demonstrated breakthrough level improvements. Research focused on using DenseNets for text classification is limited, with little to no further research conducted to determine what hyperparameters or DenseNet topologies could further improve classification performance.
\par
Designing, training and testing a deep network architecture is not a trivial task. A network topology has a learning bias, and its performance is dependant on the quality of chosen hyperparameters. The selection of these properties is based on trial and error or the practitioners' experience. Evolutionary deep learning (EDL) is an evolutionary computation technique that aims to locate performant network architectures for a particular task automatically. An evolutionary-inspired search algorithm coupled with the backpropagation \cite{backprop} algorithm is used to locate, train and evaluate a population of candidate network architectures. Each candidate architectures is encoded in a genotype, contained in the population. Each genotype is decoded to a phenotype which represents a trainable candidate architecture. EDL removes the need for expert domain knowledge and the need for a trial and error approach. To the best of our knowledge, EDL has never been used to locate candidate char-DenseNets for text classification tasks.
\par
Genetic programming (GP) \cite{gp} is an evolutionary-inspired algorithm that evolves a population of programs represented by tree structures. The task of manually constructing a neural network can be considered as a sequence of program steps taken to construct a neural network from scratch up to a performant state. These sequence of steps are the equivalent of a computer program (genotype) that when executed, constructs a neural network (phenotype). This property makes GP an appropriate algorithm to evolve neural network architectures.

\subsection{Goal}
The goal of this work is to use GP with the backpropagation algorithm to evolve performant character-level DenseNet architectures for text classification tasks automatically. This goal is achieved through the following objectives:
 \begin{enumerate}
 \item{ Introduce an appropriate indirect-encoding for representing char-DenseNet architectures that can be used in a GP-based algorithm.}
 \item{Evaluate the proposed algorithm over two well-known text classification datasets, one being of small size and the other of large size, to determine the ability of the algorithm to generalise over different text classification tasks.}
 \item{Benchmark the performance of the best-evolved models against current state-of-the-art char-CNN and char-DenseNet models.}
 \end{enumerate}
\section{Background}
\subsection{Character-level Convolutional Neural Networks}
Zhang et al. \cite{Zhang2015} demonstrated that char-CNNs are an effective approach for text classification. The approach implemented a modular design and used the back-propagation \cite{backprop} algorithm, via stochastic gradient descent \cite{8192502}, for network training. A temporal convolutional module was used for convolutional operations with their large model (Large Char-CNN) using 1024 feature maps and their small model (Small Char-CNN) using 256 feature maps.  Max-pooling allowed networks deeper than six layers. ReLU \cite{relu} was used for non-linearity, and two fully connected layers provided the classifier component of their network. Each character in the input text sequence was converted to a one-hot-vector. Each vector was then stacked to produce a matrix of vectors representing the text sequence. Text sequences were limited to a maximum of 1014 characters. Longer text sequences were truncated, and shorter sequences were padded. The authors created eight datasets to evaluate their model. The model was shallow at only six layers. Their work showed that char-CNNs perform well over large datasets but underperform on smaller datasets when compared to traditional machine learning methods.
\par
Motivated by the performance increases gained by adding depth to a network as evidenced in image classification tasks, Conneau et al. \cite{Conneau2017} introduced their \emph{very deep convolutional neural network} (VDCNN) model. Introducing the concept of a \emph{convolutional block}, consisting of a convolutional layer, a batch normalisation layer and a ReLU activation function. Their model stacked these convolutional blocks in sequential order one after the other. Further, their model implemented ResNet \cite{resnet} links to allow their model to be extended to a depth of 29 layers. Their model outperformed all current state of the art char-CNNs and demonstrated the importance of adding depth to char-CNNs. However, their model could not be extended beyond 29 layers without degrading the model's classification accuracy.
\par
Le et al. \cite{Le} conducted a study to understand the role of depth for both char-CNNs and word-CNNs. Their model was inspired by the DenseNet \cite{Huang2016} model used for image classification tasks. Using the same hyperparameters as in \cite{Zhang2015} and \cite{Conneau2017} and introducing the concept of a \emph{Dense block}, where each block consisted of multiple convolutional blocks stacked in sequential order with all convolutional blocks densely connected. Their model was able to outperform VDCNN on some datasets. Their research showed that DenseNet architectures have promising potential for text classification tasks. An interesting finding is that, again, adding depth to their model resulted in only minor improvements. The authors conclude that char-CNNs must be deep to be effective and that the problem of improving char-CNNs is not yet well understood. It is not known what other char-CNN architectures may perform better.
\subsection{Related work}
Evolutionary deep learning (EDL) is a challenging and popular research task, particularly in the image classification domain. However, there is little to no research conducted on evolving CNNs for text classification tasks. There is no research directly related to evolving char-DenseNets. One important related research work conducted was by Liang et al. \cite{Liang2019}. Their algorithm, LEAF, can evolve performant network architectures and hyperparameters concurrently for image and text classification tasks. The LEAF algorithm consists of an algorithm-layer, systems-layer and problem-domain subsystem where each component is responsible for different aspects of the algorithm. The problem-domain layer is supported by both the algorithm-layer and systems-layer. CoDeepNEAT \cite{miikkulainen2017evolving} is the underlying algorithm for the algorithm-layer, which is responsible for evolving the topology and hyperparameters of the candidate networks.
It should be noted that networks were encoded as graphs structures. CoDeepNEAT makes use of components such as LSTM, convolutional and fully connected layers, where each selected component represents a node in the network topology. Concerning text classification, their algorithm was only evaluated on the Wikipedia comment toxicity dataset \cite{Wulczyn2017}. It is noted that although the LEAF algorithm is the most well-known algorithm for evolving architectures for text classification tasks, it is not specifically designed to evolve char-CNN architectures.
\section{The Proposed Method}
The proposed method, GP-Dense, is a genetic programming-based algorithm that evolves GP trees representing executable programs that can construct trainable char-DenseNet architectures. The genotypes (GP trees) consists of executable program symbols, each representing an action to be performed in the construction of a neural network. Decoding a genotype to a phenotype involves the process of executing the program symbols in the genotype resulting in a structurally valid and trainable neural network. The overall workflow of GP-Dense is presented below in algorithm \ref{alg:evolutionary_process}.
\par
The algorithm begins by assigning a unique seed for the current experiment run. Then a population of randomly generated genotypes of varying depth is created. Each genotype is decoded to a phenotype and uploaded to the graphics processing unit (GPU). Each phenotype is trained and evaluated on the validation set to determine its fitness. The top fittest corresponding genotypes are added to the elite population. Tournament selection is used to generate a population of selected genotypes. Crossover and mutation operations are applied to the selected individuals resulting in a new population of offspring genotypes. The elite genotypes are then added to the new population. This process is repeated until a maximum number of generations has been attained, after which the fittest genotype is retrained on the full training set for an extended period of epochs and evaluated on the test set to determine its final fitness.
\\\\\\
\begin{algorithm}[H]
    \label{alg:evolutionary_process}
    \caption{GP-DenseNet.}
    \textbf{begin}\;
    \quad $seed \leftarrow$ Assign\ next\ seed\ from\ list;\\
    \quad $population \leftarrow$ genotypes\ with\ specified\ depth\ range;\\
    \While{$not\ maximum\ generations$}{%
        \ForEach{$genotype \in population$}{%
          GPU $\leftarrow phenotype \leftarrow decode(genotype)$\;
          $evaluate(genotype,\ reduced\ train.\ set,\ val.\ set)$\;
        }
        $elite \leftarrow$ fittest\ from\ population\;
        $selected \leftarrow tournament(population)$\;
        $offspring\ population \leftarrow crossover(selected)$\;
        $population \leftarrow mutate(offspring\ population)$\;
        $limit(population \cup elite)$;
     }
     $fittest \leftarrow population$\;
     $evaluate(fittest,\ full\ train.\ set,\ test\ set)$\;
    \textbf{end}\;
\end{algorithm}
\vspace{3mm}
\subsection{Encoding Network Architecture}
An appropriate genotype encoding is required to represent neural network architectures in order for evolutionary algorithms to operate on them. There are two encoding paradigms: direct and indirect encodings. Direct encoding approaches explicitly state the components and structure of the network architecture. For example, the genotype could contain convolutional layers, max-pooling and activation functions, where the position of these components is directly related to their position in the phenotype structure. A significant drawback of this approach is that evolutionary operators such as the crossover operation could result in an offspring architecture with a broken topology that is no longer a valid feed-forward neural network. Indirect encodings contain the information on how to build a neural network architecture starting from a valid base network architecture. This information can take the form of executable program symbols. For example, a program symbol can represent adding depth to the network. Evolutionary operations, such as crossover operations, have little potential to create invalid offspring networks as the selected program symbols are designed to always return a valid network topology. Indirect encodings are a natural fit for genetic programming algorithms.
\par
 Gruau et el.\cite{Gruau1994} created the well known \emph{cellular encoding} scheme for evolving multilayer perceptrons. Cellular encoding is inspired by the division of biological cells, as seen in nature. By using a set of operations on cells (nodes in the network), the encoding can represent an extensive range of network topologies from shallow to deep and narrow to wide. We propose the use of a reduced set of cellular encoding operations to evolve char-DenseNet architecture where each cell in the network represents a \emph{dense block} \cite{Le}.
  \subsubsection{Input and Classifier Layer:}
  Each character of the input text sequence is encoded as a $n$-dimensional vector
  using a lookup table that contains embeddings of a fixed alphabet. Therefore, an input text sequence of length $l$ will be encoded into a matrix of dimension $n$\ x\ $l$. The classifier layer of the ancestor network is defined as in \cite{Conneau2017}, containing a k-max pooling layer followed by three fully connected layers using ReLU \cite{relu} activation functions and softmax outputs in the final layer.
  
 \subsubsection{Dense Blocks (Cell):}
\begin{figure}[H]
\centering
\scalebox{0.85}{
\includegraphics[width=\textwidth]{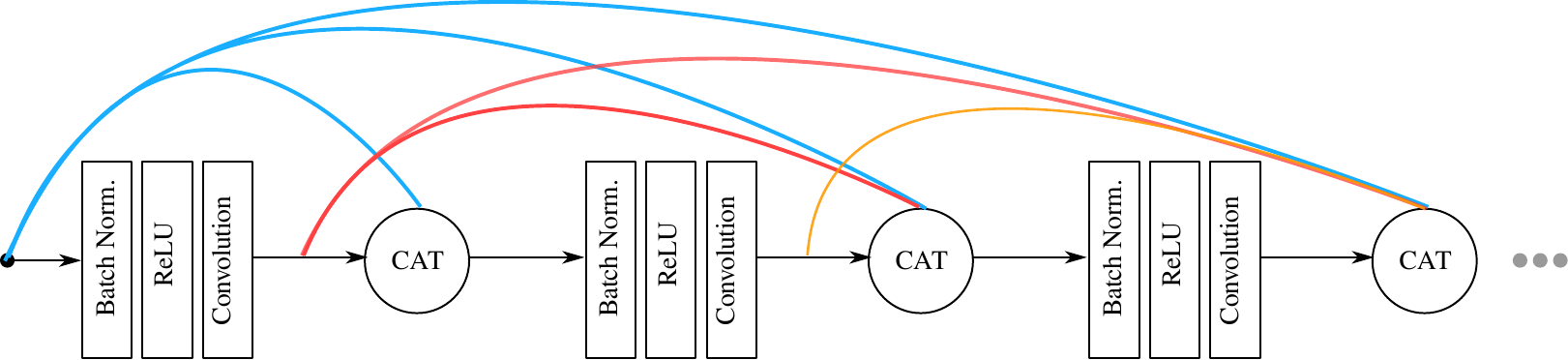}
}
\caption{Internal configuration of a dense block (cell).} \label{fig1}
\end{figure}

 A dense block (cell), as represented in figure \ref{fig1}, consists of multiple convolutional blocks as in \cite{Conneau2017}. Each convolutional block consists of a  batch normalisation layer, ReLU \cite{relu} activation function and convolutional layer. The output channels from each convolutional block are transported to the input channels of every following convolutional block in a feed-forward manner. All preceding input channels to a convolutional block are concatenated together. Densely connected blocks have been shown to increase classification performance and mitigate the vanishing gradient problem \cite{Huang2016}. Each dense block is followed by a transitional layer, as implemented in \cite{Le}, containing a 1x3 convolution and a 1x2 local max-pooling layer.
\vspace{-10mm}
 \subsubsection{GP Function Set:}
 Two operations are used from Gruau et el's.\cite{Gruau1994} cellular encoding: SEQ and PAR.
 The SEQ program symbol is a sequential split operation. The operation is provided with a \emph{mother} cell that is duplicated into a new \emph{child} cell. The mother and child cell are then connected to each other in a feed-forward sequential manner from the mother to the child. The PAR program symbol is also provided with a mother cell and produces a duplicate child cell. However, the two cells are then connected in parallel, still in a feed-forward manner. In this work, we treat a dense block as a cell. We introduce a third operation called the END operation. This operation is a \emph{no operation} symbol and is used to terminate the execution of a branch of the GP tree currently being traversed.
 \subsubsection{Terminal Set and Decorator Set:}
 Two sets are defined: \emph{Dense block parameter} terminals (DBT) and \emph{Training parameter} decorators (TPD). The DBT set consists of integers in the range from 1 to 10. This terminal set defines the number of convolutional blocks contained in a dense block (cell). DBT is used on all dense blocks (cells). The TPD consists of real values in the range 0.0 to 0.5 and represents the probability of a dense block being dropped during a batch training cycle. This function serves as a dropout feature to prevent the neural network from over-relying on a particular feature learned from the input data. If a dense block has a dropout value of 0.5, then during every second batch of training data, the dense block will be bypassed in the neural network. Each symbol in a GP tree can be \emph{decorated} with a value from the TPD terminal set to control the dense block dropout operation during the training cycle of the neural network. A probability value is used to determine if a dense block (cell) is to be decorated with the drop out feature or not. To the best of the authors' knowledge, this is the first work to include a CNN training hyperparameter in the genotype of an evolutionary algorithm.
\subsection{Decoding Network Architecture}
\begin{figure}[H]
\centering
\scalebox{0.75}{
\includegraphics[width=\textwidth]{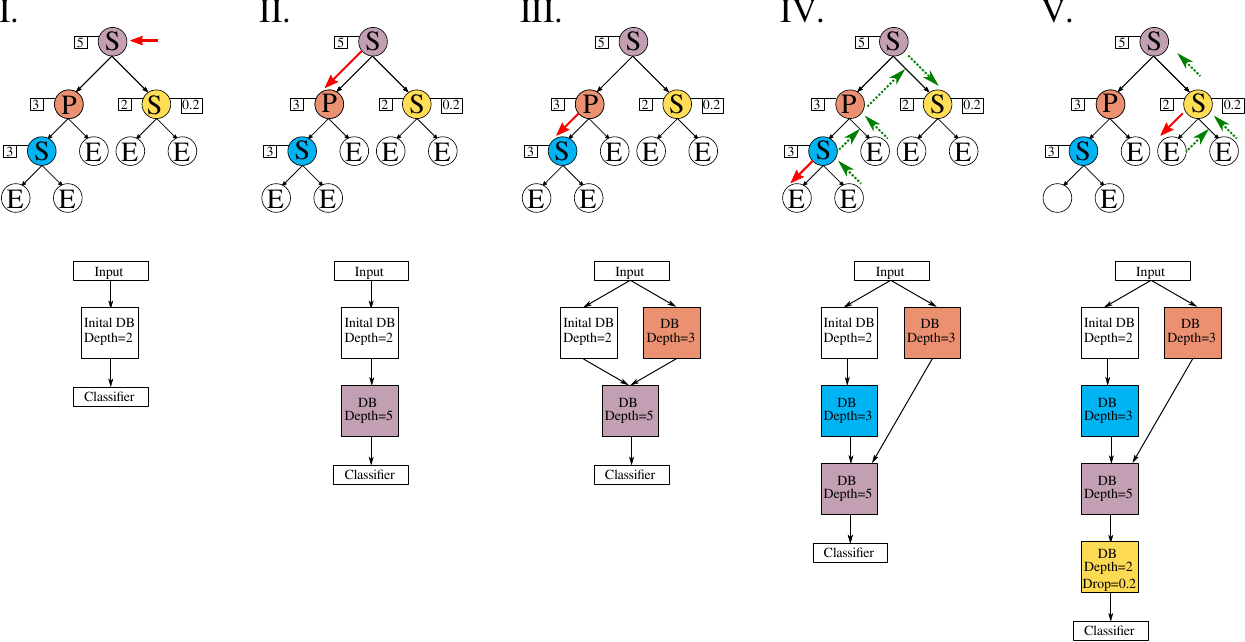}
}
\caption{Decoding genotype to phenotype. S=SEQ, P=PAR, E=END, DB=Dense Block, D=Decorator} \label{fig2}
\end{figure}
The decoding of a genotype to its corresponding phenotype is demonstrated in figure \ref{fig2}. GP-Dense uses a depth-first traversal approach when decoding a genotype. GP-Dense always start with a \emph{ancestor network} containing one dense block containing two convolutional blocks with an initial 32 input and output channels. In step I, the \emph{ancestor} network is represented by the symbol DB. The root symbol, represented by the first letter S (SEQ) of the GP tree, is executed. Step II displays the outcome. The phenotype now has two dense blocks, the original dense block and a child dense block that contains five convolutional blocks. The ancestor dense block will now be operated on by the left branch of the preceding symbol in the genotype and the child block by the right branch of the preceding genotype symbol. Step III represents the neural network after the P (PAR) symbol has been executed. Note that the PAR symbol operates on the ancestor dense block that was created by the previous genotype symbol. This ancestor dense block is now referred to as the \emph{mother} dense block and the new dense block as the child dense block. The mother and the child are now connected in parallel. The next symbols is an S (SEQ) and the output is shown in step IV. The following symbol is the END symbol which performs no operation but traces back up the GP tree as indicated by the green arrows in step IV. Step V presents the neural network after executing the S (SEQ) symbol connected to the right branch of the root cell in the genotype. This operation results in the mother dense block connected to a new child dense block, containing two convolutional blocks, in sequential order. Note that the child dense block is decorated with the real value of 0.2 , which represents a dropout function applied to the child dense block.
\subsection{Evolutionary Operators}
Tournament selection is used to select $k$ randomly selected genotypes from the population to build a breeding pool. Single point crossover is used in this work. A random point in the GP tree of each genotype is selected, and the subtrees of each genotype are exchanged at that point. This operation results in two offspring genotypes. The chosen mutation operation is uniform. A selected genotype is mutated in a random position in its GP tree structure by attaching a small randomly generated subtree at that point. The chosen crossover and mutation operations are well-researched variants and are chosen for their simplicity. Future research will focus on novel crossover and mutation operations.
\section{Experiment Design}
\subsection{Peer Competitors}
There are no related EDL algorithms with which to benchmark the proposed method. To mitigate this problem, the best evolved models from the proposed algorithm are benchmarked against current state of the hand-crafted character-level models: Le et al’s \cite{Le} char-DenseNet model, Zhang et al’s \cite{Zhang2015} Small-Char-CNN and Large-Char-CNN models and Conneau et al’s \cite{Conneau2017} VDCNN model. All the benchmark models are pure character-level models, meaning that there is no data augmentation  or text pre-processing conducted.
\subsection{ Benchmark Datasets}
Two datasets from the work of Zhang et al.\cite{Zhang2015} are selected for this research work as listed in table \ref{tab:datasets_evolution}. The AG News dataset is the smallest of the datasets selected and is regarded as a challenging dataset to classify. The second chosen dataset is the Yelp Review Full dataset which is considered a large text dataset. This dataset is particularly difficult to model as the current best accuracies level are typically still below 65\%. Each dataset is split as shown in table \ref{tab:datasets_evolution}. These values are based on the split ratio as in \cite{Zhang2015} and \cite{Conneau2017}.
\vspace{-7mm}
\begin{table}[H]
\caption{Datasets by training, validation and test splits.}\label{tab:datasets_evolution}
\centering
\begin{tabular}{lcrrr}
\hline
Dataset & No. classes & No. train & No. Validation & No. Test\\
\hline
    AG News               & 4     & 112,852 & 7,148 & 7,600 \\
    Yelp Review Full        & 5     & 603,571   & 46,429 & 50,000 \\
\hline
\end{tabular}
\end{table}
\vspace{-4mm}
\noindent
Table \ref{tab:datasets_stats} lists the statistics of the instances in each dataset. The AG News dataset consists of text sentences of 256 characters on average.  Zhang et al.’s \cite{Zhang2015}, Conneau et al's.\cite {Conneau2017} and Le et al's. \cite{Le} models all use a temporal length of 1014  characters. This implies redundant padding and wasted convolutional operations. Therefore, this work  uses a maximum sentence length of 256 characters when evolving models for the AG news dataset and 512 characters for the Yelp Reviews Full dataset.
\vspace{-10mm}
\begin{table}[H]
\caption{Sentence statistics.}\label{tab:datasets_stats}
\centering
\begin{tabular}{lrrr}
\hline
Dataset & Mean & Minimum & Maximum\\
\hline
    AG News  & 236$\pm$66 & 100 & 1,012 \\
    Yelp Review Full & 732$\pm$664 & 10 & 5,849\\
\hline
\end{tabular}
\end{table}
\subsection{Parameter Settings}
\vspace{-10mm}

\begin{table}[H]
  \caption{Parameter settings.}
  \centering
  \scalebox{0.75}{
    \begin{tabular}{lr}
   \hline
    Parameter                 & Value\\
    \hline
    Run count, seed             & 30,  Unique per Run \\
    Epochs, batch size     & 10, 128 \cite{Zhang2015},\cite{Conneau2017} \\
    Initial Learning Rate, momentum    & 0.01, 0.9 \cite{Conneau2017} \\
    Learning Schedule & Halve every 3 epochs \cite{Zhang2015} \\   
    Weight Initialisation & Kaiming\cite{Kaiming2018a}\cite{Zhang2015},\cite{Conneau2017}\\
    Training Data Usage & 0.25\\
    Alphabet & Same as in \cite{Zhang2015} and \cite{Conneau2017} \\   
    Max Sentence Length & 256 (AG News) 512 (Yelp)\\
    Kernel size, stride, padding & 3,1,1 \cite{Conneau2017} \\
    Number of Generations & 20  \\
    Population Size, elitism size & 20, 0.1\\
    Crossover Prob and type,  Mut. Prob. and type & 0.5 Single, 0.1, Unif. \\
    Mutation Growth, size & Grow, [1,3] \\
    Tournament Selection Size & 3 \\
    Initial, max Tree Depth & [1,10], 17 \cite{gp} \\
    Fitness Function & max(val. acc.) \cite{Zhang2015} \cite{Conneau2017}\\ 
    Probability Dropout applied to cell & 0.1\\
    \hline
  \end{tabular}
  }
  \label{tab:parameters}
\end{table}
Table \ref{tab:parameters} lists the parameter settings for the experiment. Thirty runs of the algorithm are conducted for each dataset. Each run is assigned a single unique seed. Candidate architectures are trained for 10 epochs to mitigate the problem of long training times. The values for the learning rate, momentum and batch size are the same as in \cite{Zhang2015} and \cite{Conneau2017}. Stochastic gradient descent is used for the network optimiser as used in \cite{Zhang2015}. Each population run consists of 20 individuals and is evolved over 20 generations. Elitism, crossover and mutation rate settings are values found in the literature \cite{gp}. The mutation growth depth is set to a low value to ensure that only small changes are made to the genotype so as not to significantly alter the phenotype. The maximum tree depth is set to a value of 17 and is considered best practice \cite{gp}. The objective is to maximise the candidate architectures validation accuracy.
\section{Results and Discussions}
\vspace{-10mm}
\begin{figure}[H]
\centering
\includegraphics[width=\textwidth]{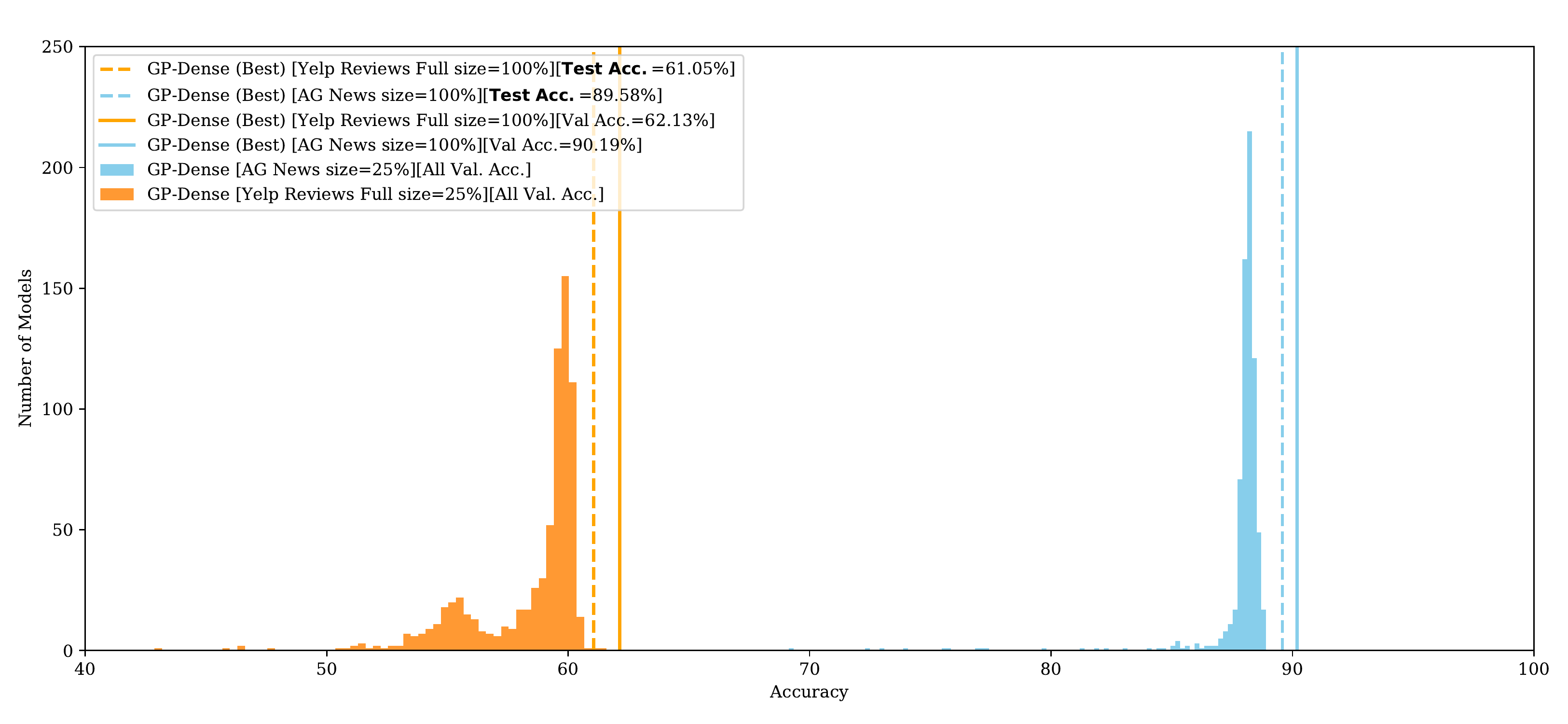}
\caption{Distribution of validation accuracy for evolved architectures.}
\label{fig:histogram}
\end{figure}
\vspace{-5mm}
The validation accuracies of all the models evolved by the GP-Dense algorithm over the AG News and Yelp Reviews Full dataset are presented in figure \ref{fig:histogram}. The distribution in orange represents the validation accuracies attained over the Yelp Reviews Full dataset, and the distribution in blue represents those attained over the AG News dataset. It is noted that these validation accuracies are attained using only 25\% of each dataset. Both distributions are left-skewed, indicating that GP-Dense has managed to maintain populations of models with similar good validation accuracies and only a few of lower quality. It can be observed that GP-Dense attained a maximum validation accuracy of 62.13\% for the Yelp Reviews Full dataset and 90.19\% for the AG News dataset. These values are achieved when retraining the best-evolved models for each dataset using 100\% of the training set. Evaluating each of the best fully trained models yields test accuracies of 61.05\% for Yelp Reviews Full and 89.59\% for the AG News dataset.
\par
\begin{figure}[H]
  \centering
  \scalebox{1.0}{
\subfloat[AG News.]{
  \hspace{-7mm}
\scalebox{0.56}{
  \includegraphics[width=1.0\linewidth]{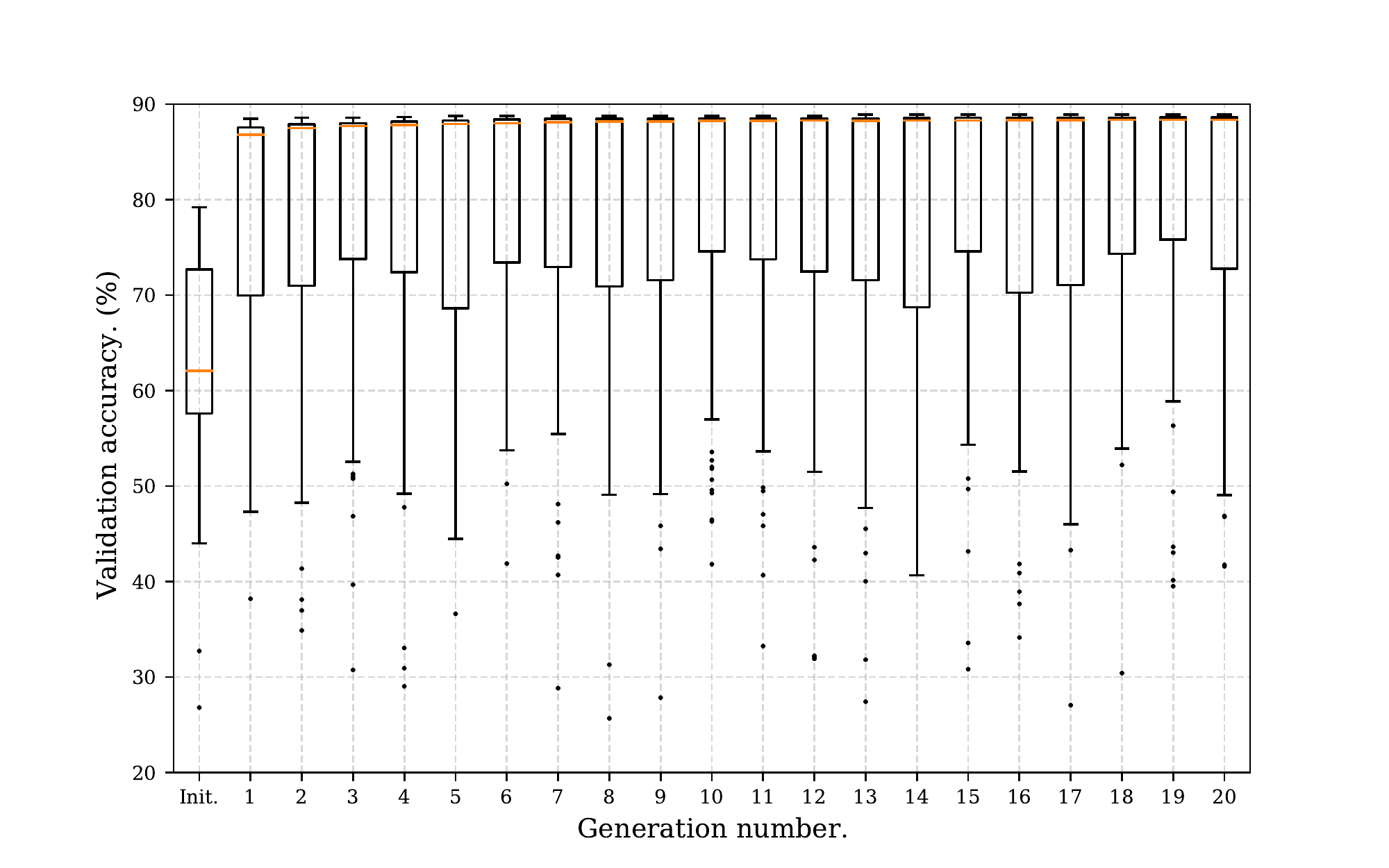}
 \label{fig:generations_a}
  }}
\subfloat[Yelp Reviews Full.]{
\hspace{-10mm}
\scalebox{0.56}{
  \includegraphics[width=1.0\linewidth]{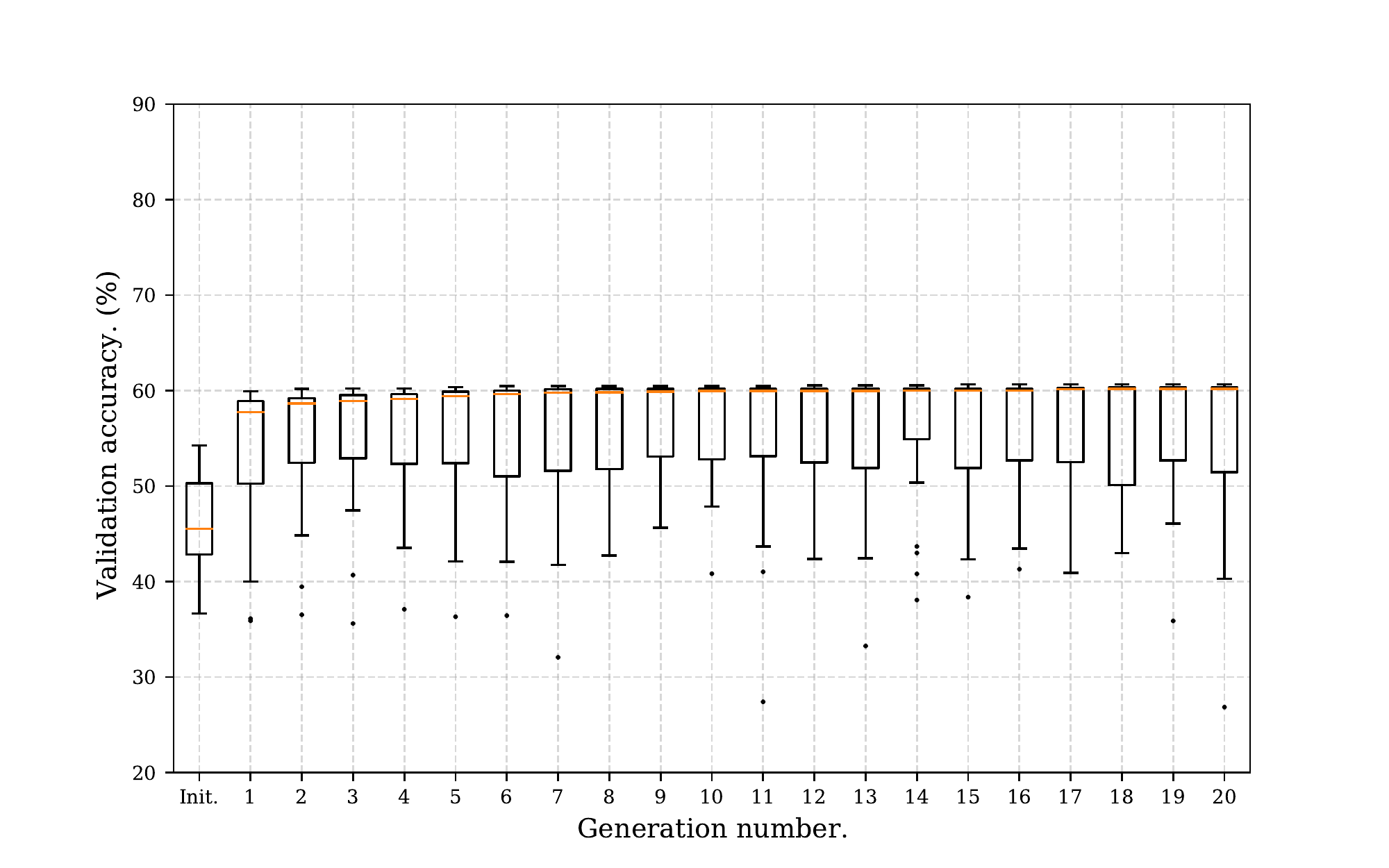}
   \label{fig:generations_b}
    }
  }
  }
\caption{GP-Dense performance over generations.}
\label{fig:generations}
\end{figure}
\begin{figure}[H]
  \centering
  \scalebox{0.85}{
\subfloat[AG News.]{
  \hspace{-9mm}
\scalebox{0.67}{
  \includegraphics[width=0.9\linewidth]{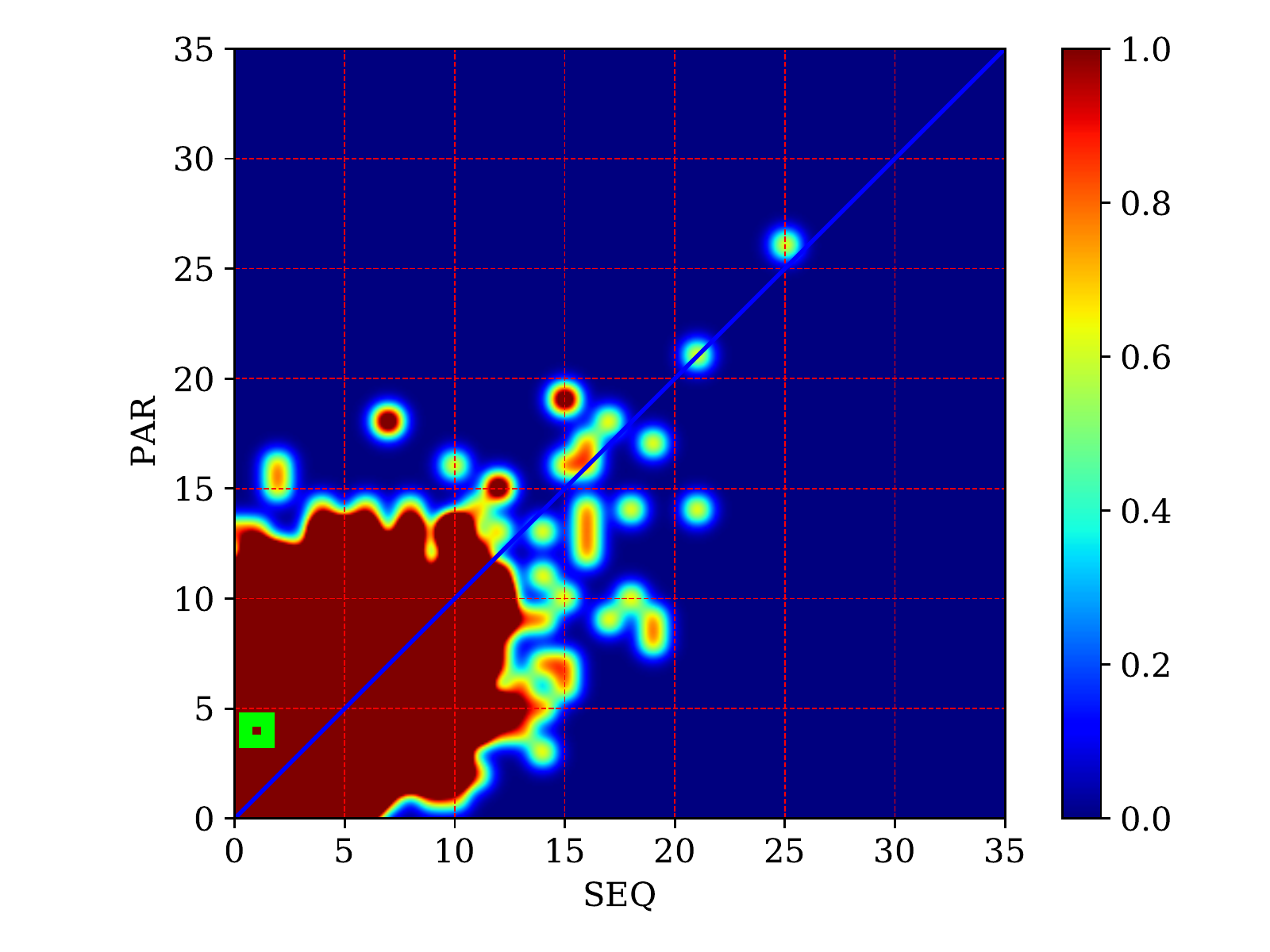}
 \label{fig:heatmap_a}
  }}
\subfloat[Yelp Reviews Full.]{
\hspace{-9mm}
\scalebox{0.67}{
  \includegraphics[width=0.9\linewidth]{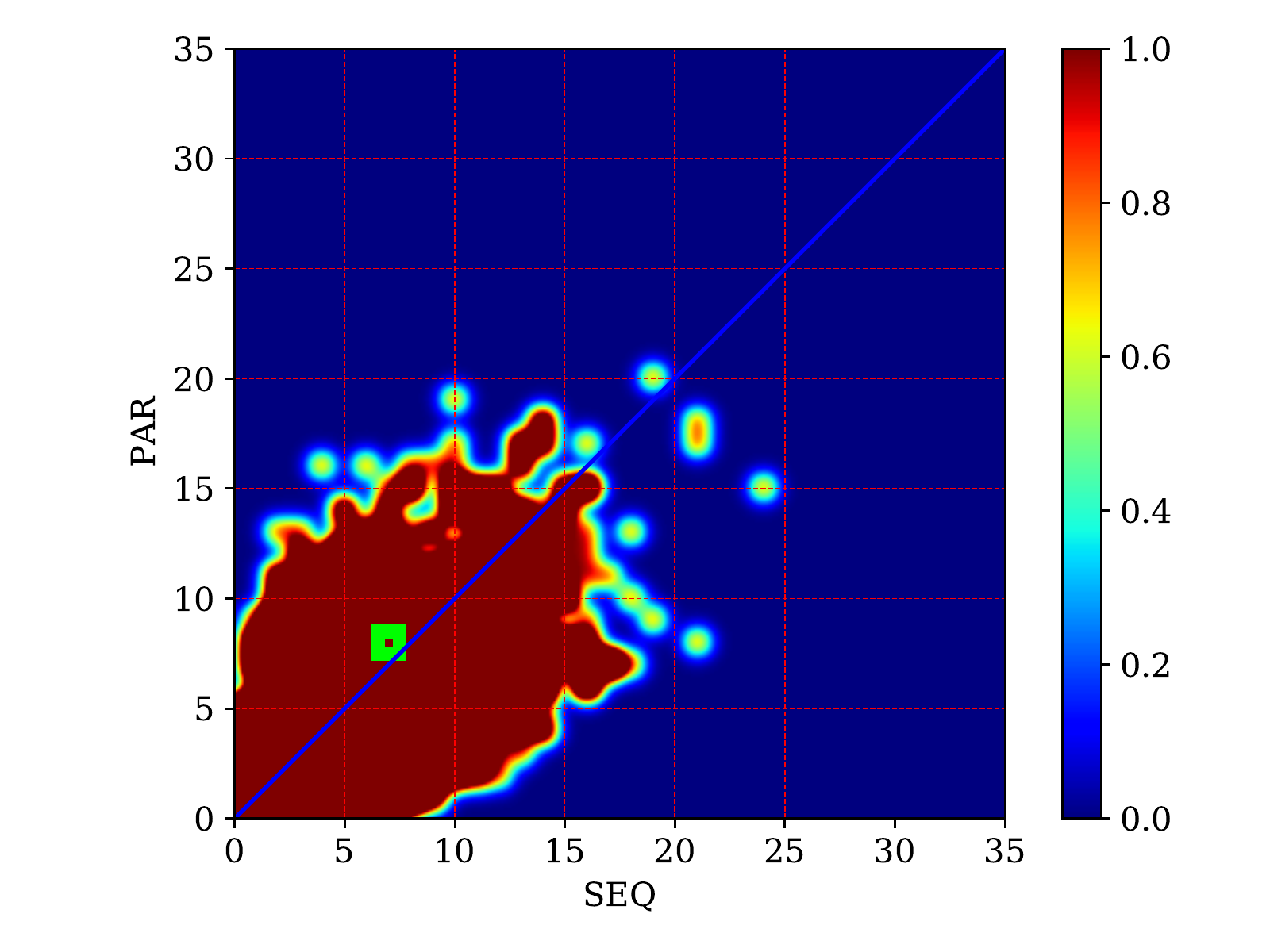}
   \label{fig:heatmap_b}
    }
  }
  }
\caption{Density of number of SEQ vs PAR operations.}
\label{fig:density_heatmap}
\end{figure}
\begin{figure}[H]
\centering
\scalebox{0.85}{
\includegraphics[width=\textwidth]{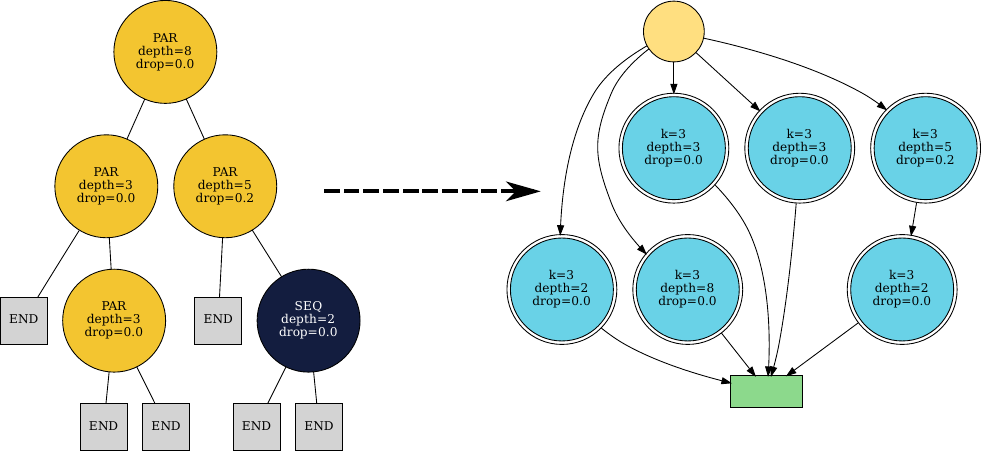}
}
\caption{AG News: Best evolved genotype and corresponding phenotype.}
\label{fig:fittest_ag_news}
\end{figure}
\begin{figure}[H]
\centering
\scalebox{1.0}{
\includegraphics[width=\textwidth]{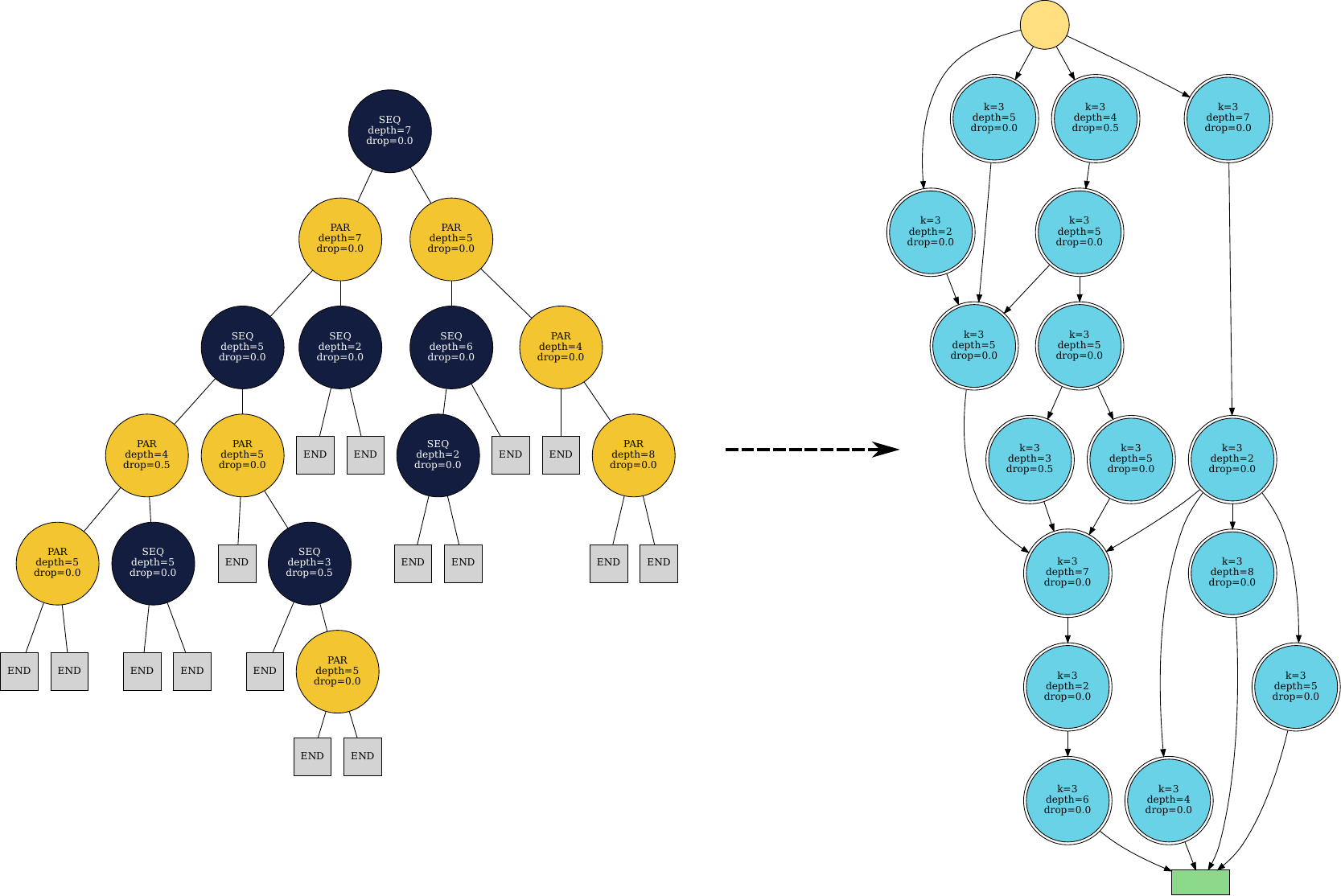}
}
\caption{Yelp Reviews Full: Best evolved genotype and corresponding phenotype.}
\label{fig:fittest_yelp}
\end{figure}

Figure \ref{fig:generations} presents the combined validation performance of all evolved models over each generation for 30 runs. GP-Dense converges to a maximum validation accuracy early in the evolutionary process and maintains this position through to the end of the evolutionary process. There are scatterings of low-quality models that appear in each generation. An analysis of these models indicates that they are the result of crossover and mutation operations that resulted in low-performance models.
\vspace{-7mm}
\vspace{-5mm}
\noindent
The distribution of the number of SEQ and PAR operations that are contained in the genotypes of all evolved models is presented in figure \ref{fig:density_heatmap}. There is no evidence that the evolutionary process favoured either operation over the other. Each of the best-evolved models is highlighted by a lime green square. The best-evolved model from AG News contained 4 PAR operations and 1 SEQ operation in its genotype, indicating that the phenotype is wide and shallow. The best model evolved for Yelp Reviews Full contained 7 PAR operations and 8 SEQ operations, indicating that the phenotype is likely to be deep and relatively wide.
\vspace{-5mm}
\noindent
The fittest genotype and corresponding phenotype evolved for the AG News dataset is presented in figure 
\ref{fig:fittest_ag_news}. Note that the phenotype has eight dense blocks connected in parallel. It can also be observed that the deepest path in the network is eight convolutional layers deep. The networks depth is comparable to Zhang et al.'s \cite{Zhang2015} original char-CNN network. The fittest genotype and corresponding phenotype for the Yelp Reviews Full dataset set are presented in figure \ref{fig:fittest_yelp}. This network is significantly deeper at 34 convolutional layers deep. The network is comparable to the depth of VDCNN \cite{Conneau2017} at 29 convolutional layers.
\newpage
\begin{figure}[H]
\centering
\scalebox{0.9}{
\includegraphics[width=\textwidth]{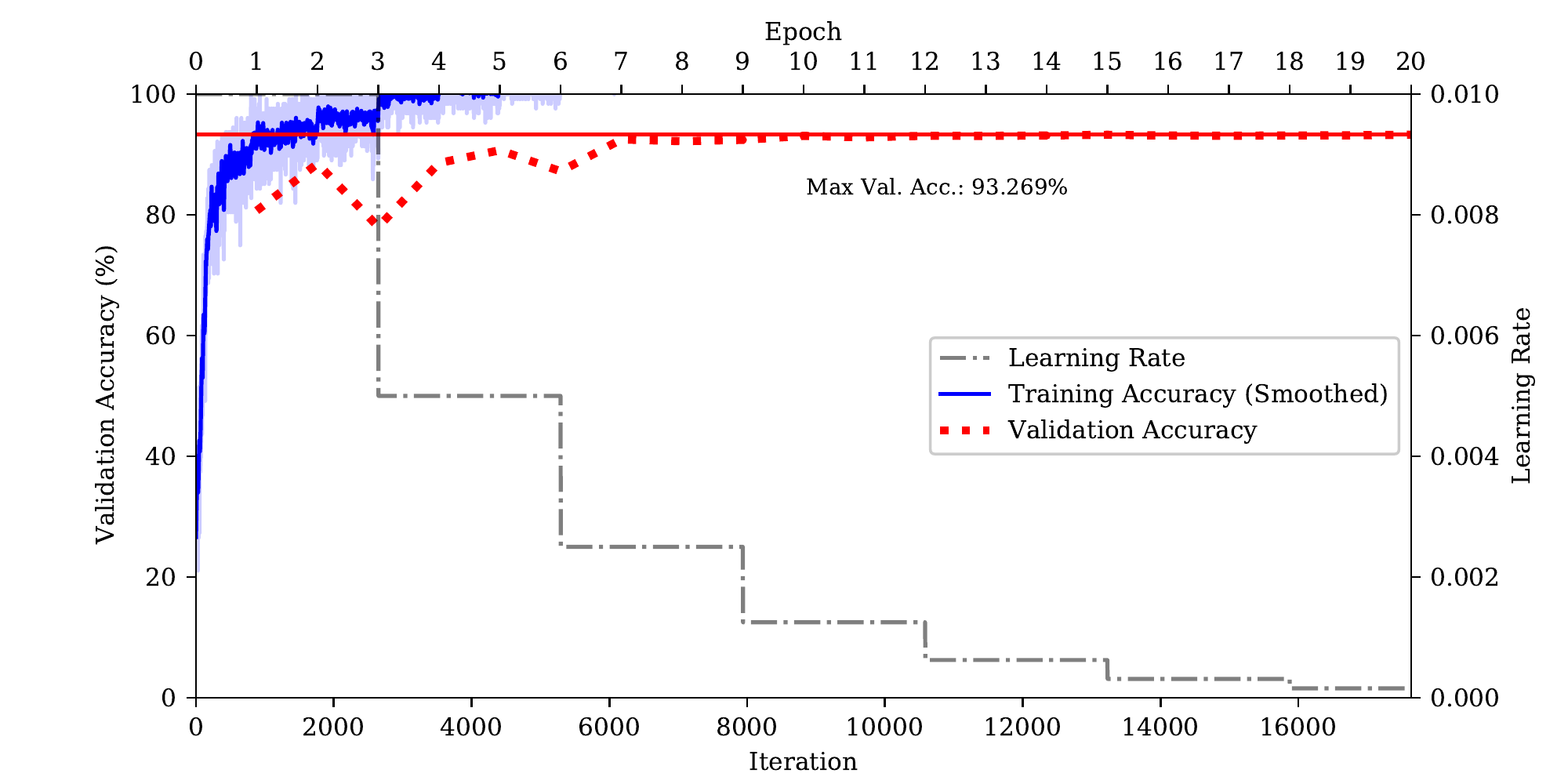}
}
\caption{Training response of best evolved model for AG News dataset.}
\label{fig:training_ag_news}
\end{figure}
\vspace{-5mm}
\noindent
The training response for the fittest phenotype evolved over the AG News dataset is presented in figure \ref{fig:training_ag_news}. The model converged within six epochs, highlighting the evolutionary pressure applied by GP-Dense to evolve models that converge within ten epochs. The validation accuracy degraded after the third epoch. However, after halving the learning rate, the validation accuracy increased again. The same behaviour is seen when the learning rate halved again at the sixth epoch. Further reducing the learning rate stabilised the validation accuracy.
\vspace{-8mm}
\begin{figure}[H]
\centering
\scalebox{0.9}{
\includegraphics[width=\textwidth]{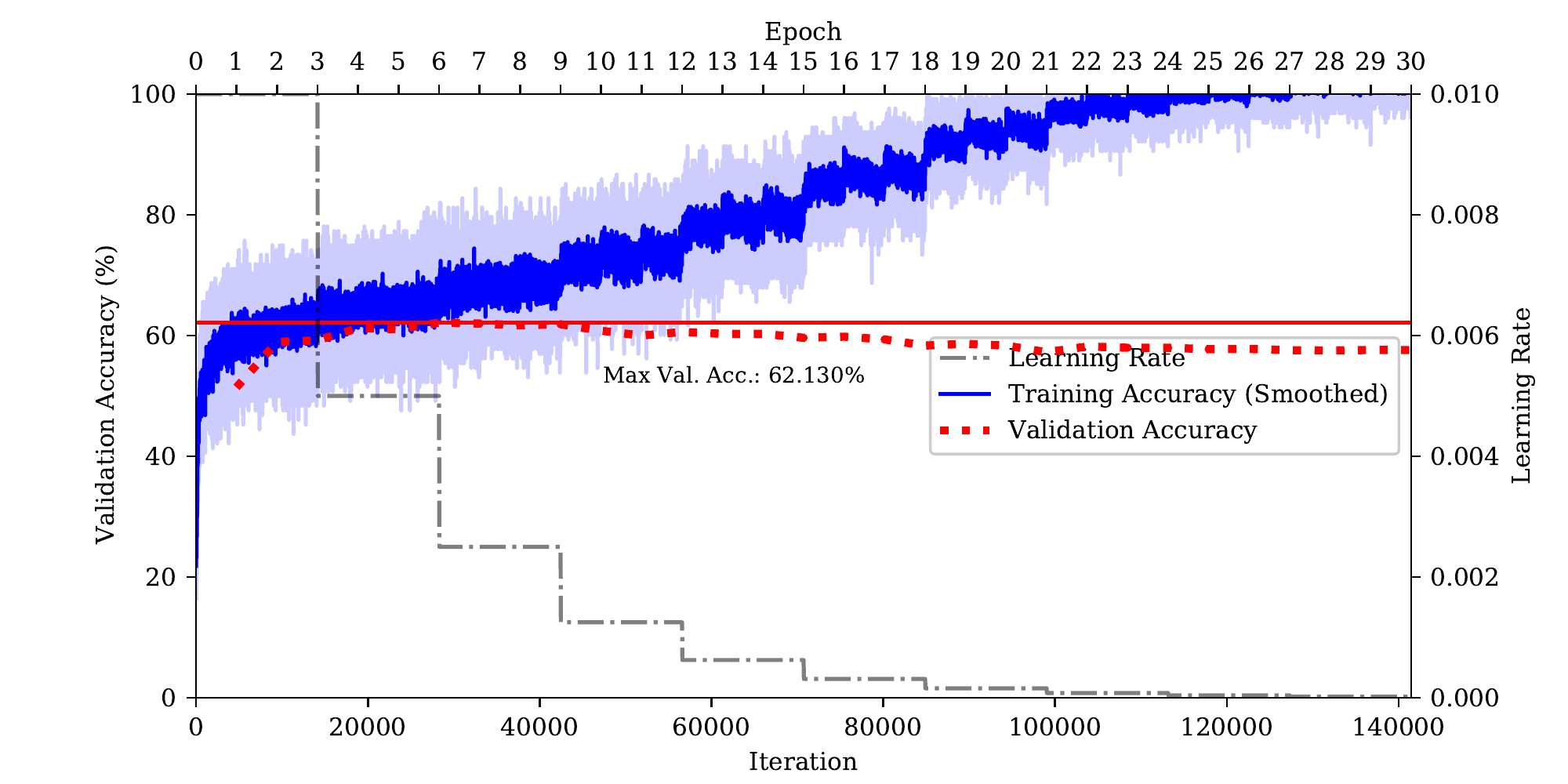}
}
\caption{Training response of best evolved model for Yelp Reviews Full dataset.}
\label{fig:training_yelp}
\end{figure}
\noindent
Figure \ref{fig:training_yelp} presents the training response of the fittest evolved phenotype over the Yelp Reviews Full dataset. Halving the learning rate every three epochs did not provide any benefit in improving the accuracy of the model. The model began overfitting after the tenth epoch. This observation suggests that GP-Dense may need to evolve a candidate model for longer than ten epochs when training on large datasets. This approach would decrease the probability of models overfitting in later epochs.
\vspace{-5mm}
\begin{table}[H]
    \centering
    \caption{Test accuracies attained.}
    \begin{tabular}{lrrrr}
        \hline
        Model or Algorithm  & AG News & Params. & Yelp Reviews & Params.\\
        \hline
        Large Char-CNN (Zhang et al. \cite{Zhang2015}) & 87.18 & \textasciitilde15&60.38&\textasciitilde15\\
        Small Char-CNN (Zhang et al. \cite{Zhang2015}) &  \textcolor{red}{\textbf{84.35}} &  \textasciitilde11&\textcolor{red}{\textbf{59.16}}&\textasciitilde11\\
        VDCNN-29 (Conneau et al. \cite{Conneau2017}) &91.27 & \textcolor{red}{\textbf{\textasciitilde17}} & \textcolor{blue}{\textbf{64.72}} & \textcolor{red}{\textbf{\textasciitilde17}}\\
        char-DenseNet (Le et al.) \cite{Le} &  \textcolor{blue}{\textbf{92.10}}  & -& 64.10  & -\\
        \textbf{GP-Dense Best (ours)} & 89.58 & \textcolor{blue}{\textbf{\textasciitilde4}}&61.05&\textcolor{blue}{\textbf{\textasciitilde7}}\\
        \hline
    \end{tabular}
    \label{tab:modelResults}
\end{table}
\noindent
The comparison of test accuracies with the other four hand-crafted state-of-the-art models are presented in table \ref{tab:modelResults}. GP-Dense outperformed all char-CNN models by Zhang et al. \cite{Zhang2015} for both the AG News and Yelp Reviews Full dataset. The evolved model underperformed VDCNN \cite{Conneau2017} by an absolute value of 1.69\% on AG News and 3.67\% for Yelp Reviews Full. However, VDCNN has almost triple the number of trainable parameters than the model evolved by GP-Dense on AG News more than double on Yelp Full Reviews. char-DenseNet outperformed GP-Dense by an absolute value of 2.52\% on AG News and 3.05\% on Yelp Reviews Full. The parameter size for char-DenseNet is not reported in \cite{Le}. GP-Dense evolved models for both AG News and Yelp Reviews Full that are smaller than all other reported models.
\section{Conclusions}
This work proposed an evolutionary deep learning algorithm (GP-Dense) to evolve character-level DenseNet architectures for text classification tasks automatically. This goal was achieved through the implementation of a GP-based evolutionary algorithm using an indirect encoding to represent candidate neural network architectures. GP-Dense was able to successfully evolve competitive network architectures using only 25\% of the datasets involved. It was observed that GP-Dense performed better on the smaller dataset, AG News, than on the larger dataset, Yelp Reviews Full. EDL has the potential to perform well on evolving character-level DenseNet architectures for text classification tasks as evidenced by this work. However, it has demonstrated that further research such as parameters tuning and using a larger subset of a training dataset is required to ensure that performant architectures can be evolved for larger datasets.


%
%
%
\bibliographystyle{splncs03_unsrt}
\bibliography{ref.bib}
%




\end{document}